%% file: Formatting-Instructions-LaTeX-2026.tex
\documentclass[letterpaper]{article} 
\usepackage{aaai2026}  
\usepackage{times}  
\usepackage{helvet}  
\usepackage{courier}  
\usepackage[hyphens]{url}  
\usepackage{graphicx} 
\usepackage{natbib}  
\usepackage{caption} 
\usepackage{algorithm}
\usepackage{algorithmic}

\usepackage{newfloat}
\usepackage{amssymb}
\usepackage{listings}
\DeclareCaptionStyle{ruled}{labelfont=normalfont,labelsep=colon,strut=off} 
\floatstyle{ruled}
\newfloat{listing}{tb}{lst}{}
\floatname{listing}{Listing}
\title{VirtualEnv: A Platform for Embodied AI Research}

\author{
  Kabir Swain\textsuperscript{\rm 1},
  Sijie Han\textsuperscript{\rm 2},
  Ayush Raina\textsuperscript{\rm 3},
  Jin Zhang\textsuperscript{\rm 3},
  Shuang Li\textsuperscript{\rm 1},
  Michael Stopa\textsuperscript{\rm 3},
  Antonio Torralba\textsuperscript{\rm 1}
}

\affiliations{
  \textsuperscript{\rm 1}Massachusetts Institute of Technology\\
  \textsuperscript{\rm 2}University of Toronto\\
  \textsuperscript{\rm 3}Sony Interactive Entertainment\\
  kswain@mit.edu,
  hs.han@mail.utoronto.ca,
  ayush.raina@sony.com,
  jin.1.zhang@sony.com,
  lishuang@mit.edu,
  michael.stopa@sony.com,
  torralba@mit.edu
}

\usepackage{bibentry}

\usepackage{subcaption}
\usepackage{graphicx}
\usepackage{minted}
\usepackage{tabularx}
\usepackage{enumitem}
\usepackage{dblfloatfix}
\usepackage{booktabs}
\usepackage[all]{nowidow}
\usepackage{float}

\begin{document}

\maketitle

\input{text/abstract}
\input{text/introduction}

\input{text/related-work}

\input{text/environment}

\input{text/experiments}

\input{text/conclusion}
\input{text/acknowledgments}

\bibliography{main.bib}

\end{document}

%% file: text/abstract.tex
\begin{abstract}

\begin{figure*}[h]
    \hfill
    \includegraphics[width=\linewidth]{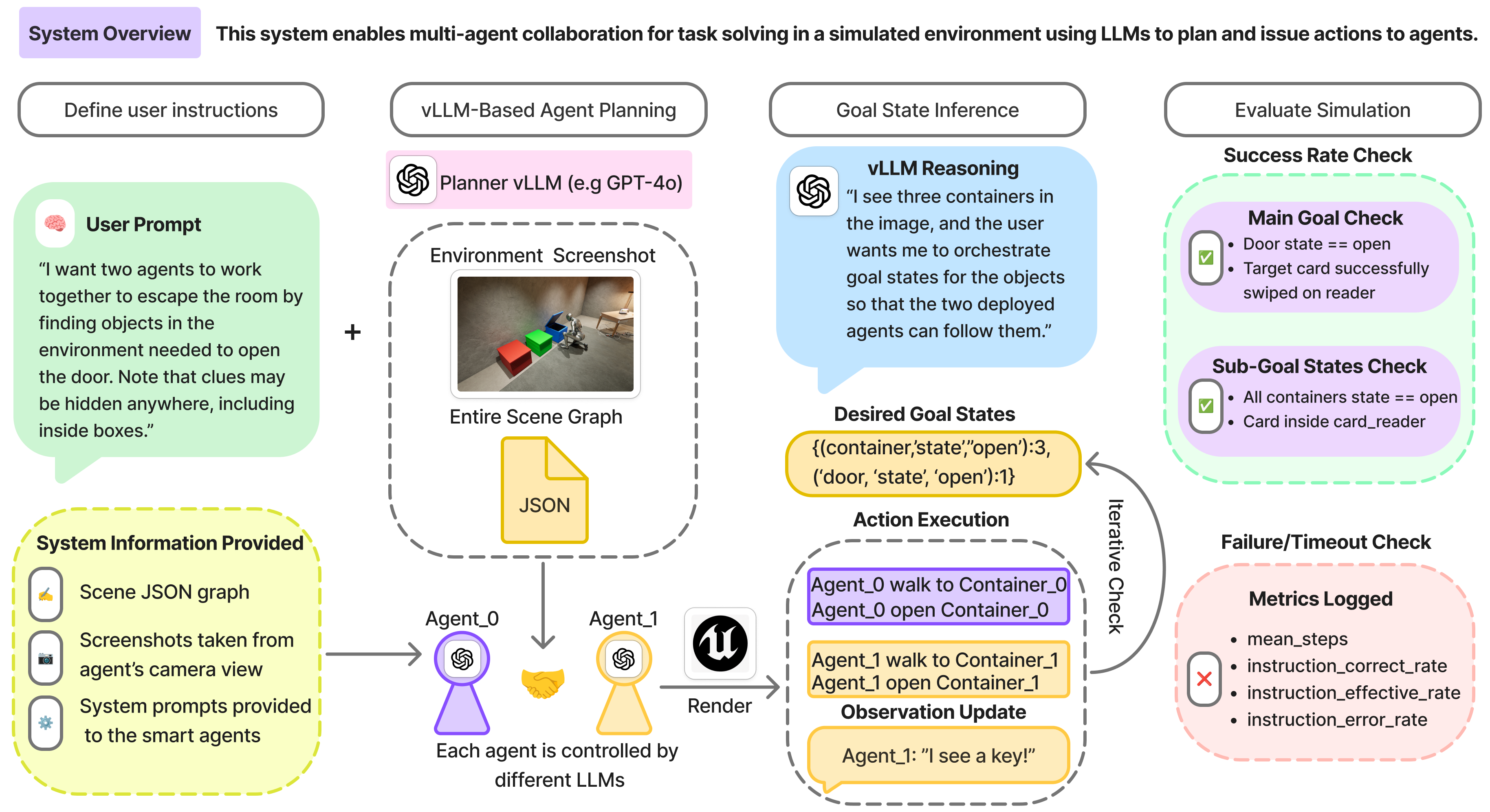}
    \caption{System overview of multi-agent planning and execution in VirtualEnv. VirtualEnv is a high-fidelity simulation environment built on Unreal Engine 5, designed for evaluating large language models (LLMs) in interactive, task-oriented settings. This figure shows the full pipeline for multi-agent task execution: users provide natural language instructions, and the system uses vLLMs (e.g., GPT-4o) to interpret goals, generate symbolic plans, and coordinate multiple agents in real time. Each agent receives environment information—including a scene graph and visual context—and executes actions via the VirtualEnv API. Performance is evaluated through goal completion checks and instruction-based success metrics.}
    \label{fig:system_overview}
\end{figure*}

As large language models (LLMs) continue to improve in reasoning and decision-making, there is a growing need for realistic and interactive environments where their abilities can be rigorously evaluated. We present VirtualEnv, a next-generation simulation platform built on Unreal Engine 5 that enables fine-grained benchmarking of LLMs in embodied and interactive scenarios. VirtualEnv supports rich agent–environment interactions, including object manipulation, navigation, and adaptive multi-agent collaboration, as well as game-inspired mechanics like escape rooms and procedurally generated environments. We provide a user-friendly API built on top of Unreal Engine, allowing researchers to deploy and control LLM-driven agents using natural language instructions. We integrate large-scale LLMs and vision-language models (VLMs), such as GPT-based models, to generate novel environments and structured tasks from multimodal inputs. Our experiments benchmark the performance of several popular LLMs across tasks of increasing complexity, analyzing differences in adaptability, planning, and multi-agent coordination. We also describe our methodology for procedural task generation, task validation, and real-time environment control. VirtualEnv is released as an open-source platform, we aim to advance research at the intersection of AI and gaming, enable standardized evaluation of LLMs in embodied AI settings, and pave the way for future developments in immersive simulations and interactive entertainment.


\end{abstract}

%% file: text/introduction.tex
\section{Introduction}

Simulators have become essential tools for researchers to develop, evaluate, and test AI models in controlled, reproducible, and scalable environments \citep{DBLP:journals/corr/abs-1904-01201, DBLP:journals/corr/abs-1712-05474, xia2018gibson}. They offer a cost-effective way to generate large-scale data and enable the study of complex models such as deep neural networks. Simulators are widely used across computer vision \cite{muller2018sim4cv, 9636667, li2024behavior1khumancenteredembodiedai, ge2024behavior, li2024photorealistic, rudek2024building} and reinforcement learning \cite{ferigo2020gym, dosovitskiy2017carla, brockman2016openai, tassa2018deepmind, kaup2024review} for tasks such as scene understanding, robotic navigation, and object interaction. In parallel, simulation platforms have increasingly been adopted in gaming research for character control \cite{parberry2017introduction, todorov2012mujoco} and procedural gameplay mechanics \cite{bel2011developing, interactive2016hitman, perez2015dynamic}.

Despite these advances, existing simulators remain limited in scale, diversity, and interactivity. Many focus exclusively on small indoor household settings (e.g., VirtualHome \cite{DBLP:journals/corr/abs-1806-07011}, House3D \cite{wu2018building}, AI2-THOR \cite{DBLP:journals/corr/abs-1712-05474}), with rigid environments and static object arrangements. These constraints hinder progress on tasks requiring generalization, planning, and emergent behavior. Simulators designed for gaming typically offer higher visual fidelity but often lack the modularity, programmability, and semantic richness required for embodied AI research. As research increasingly incorporates large language models (LLMs) and vision-language models (VLMs) into embodied settings, there is a growing need for flexible simulation platforms that support multimodal grounding, interactive task generation, and dynamic environment editing at scale.

To address these limitations, we introduce \textbf{VirtualEnv}, a next-generation simulation platform built on Unreal Engine 5 \cite{unrealengine5}, designed to support language-driven and multimodal research in embodied AI. VirtualEnv offers a modular framework for simulating expansive, richly interactive environments that span urban settings, multi-room buildings, and outdoor spaces. The platform supports fine-grained agent-object interactions, dynamic scene editing, and procedural environment generation through integration with LLMs and VLMs (See Figure~\ref{fig:system_overview}). Unlike prior simulators, VirtualEnv enables real-time interaction with large-scale environments and supports a wide range of tasks, including spatial reasoning, tool use, goal-conditioned planning, and multi-agent collaboration.

To demonstrate its capabilities, we introduce a suite of Escape Room–style environments \cite{Heikkinen2016DesigningAE}, where LLM-driven agents are tasked with solving cognitive puzzles that require multi-step reasoning, object manipulation, and sequential planning. These scenarios serve as benchmarks for evaluating language models in grounded, task-oriented environments with escalating difficulty levels. Our experiments compare several vLLMs across task success rate, instruction following, and generalization under environmental variation.

By releasing VirtualEnv as an open-source platform, we aim to provide the community with a standardized testbed for multimodal learning, embodied reasoning, and simulation-driven AI development. In particular, we position VirtualEnv as a foundation for benchmarking large language models in interactive environments—enabling consistent, reproducible comparisons across tasks, modalities, and levels of embodiment. We hope VirtualEnv accelerates progress at the intersection of AI, gaming, and simulation, offering a flexible foundation for research in language-guided agents, procedural task generation, and virtual environment control.

%% file: text/related-work.tex

\section{Related Work}
\begin{figure*}[!t]
    \hfill
    \includegraphics[width=\linewidth]{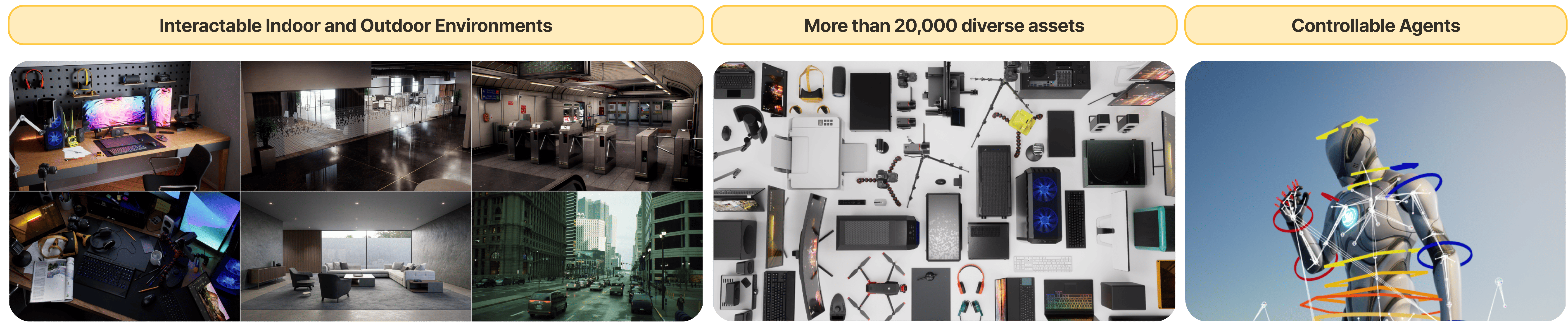}
    \caption{Core capabilities of VirtualEnv. The platform supports highly realistic and interactable indoor and outdoor environments, a large curated library of over 20{,}000 diverse assets, and controllable humanoid agents with fine-grained motion support. These features enable the creation of complex, multimodal simulation scenarios suitable for embodied AI training, evaluation, and benchmarking.}

    \label{fig:example}
\end{figure*}

\begin{table*}[!t]
    \centering
    \renewcommand{\arraystretch}{1.2}
    \resizebox{\textwidth}{!}{%
    \begin{tabular}{l c c c c c r}
        \toprule
        & \textbf{Environment} & \textbf{Multi-Agent} & \textbf{Language} & \textbf{Action Space} & \textbf{Task Types} & \textbf{Num tasks} \\
        \midrule
        AI2Thor (Kolve et al., 2019) & 3D-S & & \checkmark & HL & CST & 48,000 \\
        OmniGibson (Li et al., 2023a) & 3D-M & & \checkmark & LL+HL & CST & 1,000 \\
        VirtualHome (Puig et al., 2021) & 3D-M & \checkmark & & HL & C & 1,200 \\
        Habitat 3.0 (Puig et al., 2023) & 3D-M & \checkmark & \checkmark & LL+HL & CSTH & 100,000 \\
        \midrule
        \textbf{VirtualEnv} & 3D-MIO & \checkmark & \checkmark & HL & CSTH & 140,000 \\
        \bottomrule
    \end{tabular}%
    }
    \caption{Overview of Embodied AI simulation platforms and their key features. We position \textbf{VirtualEnv} alongside widely used simulators across several dimensions: scene complexity (single-room (S), multi-room (M), indoor-outdoor (IO)), support for multiple agents, language interaction, and action space (high-level (HL), low-level (LL)). Task coverage is annotated using the following categories: constraint-free (C), spatial (S), temporal (T), and heterogeneous (H). "Num tasks" reflects the approximate number of unique, predefined or generated scenarios supported by each platform.}
    
    \label{tab:comparison}
\end{table*}

Simulators have become indispensable tools in computer vision, embodied AI, and reinforcement learning research, serving as environments for data generation, algorithm validation, and performance benchmarking. Several platforms have been developed, each tailored to specific research goals. Popular examples include VirtualHome \cite{DBLP:journals/corr/abs-1806-07011}, AI2-THOR \cite{DBLP:journals/corr/abs-1712-05474}, OmniGibson \cite{li2024behavior1khumancenteredembodiedai}, Habitat \cite{DBLP:journals/corr/abs-1904-01201}, ProcTHOR \cite{procthor}, UnrealCV \cite{qiu2016unrealcv}, and UnrealZoo \cite{zhong2025unrealzoo}. While these simulators exhibit varying degrees of interactivity, realism, and complexity, they typically emphasize specific capabilities such as navigation, manipulation, or visual perception, limiting their applicability to broader AI research tasks.

\textbf{VirtualHome} \cite{DBLP:journals/corr/abs-1806-07011} is designed explicitly to simulate household environments, providing structured scene graphs and scripted activities for everyday tasks such as cooking and cleaning. Although influential in facilitating high-level reasoning and task planning, VirtualHome is constrained by its exclusive focus on indoor residential settings, limiting the scope and diversity of potential research scenarios.

\textbf{AI2-THOR} \cite{DBLP:journals/corr/abs-1712-05474} expands the range of embodied AI research through interactive indoor scenarios that support visual question answering, navigation, and object manipulation. Despite its extensive adoption, AI2-THOR similarly emphasizes indoor environments, lacking the broader urban or outdoor contexts necessary for more generalizable AI agent development.

\textbf{OmniGibson} \cite{li2024behavior1khumancenteredembodiedai} extends interactive simulations by including dynamic object states, physics-driven interactions, and virtual reality interfaces to facilitate transfer to real-world settings. Its detailed indoor scenarios have proven useful for navigation and manipulation tasks. Nevertheless, its emphasis remains largely domestic, limiting applicability to complex urban environments and large-scale AI tasks involving diverse interactions.

\textbf{Habitat} \cite{DBLP:journals/corr/abs-1904-01201} provides a high-performance engine primarily optimized for navigation and exploration, enabling scalable training and evaluation of embodied agents. However, Habitat's primary limitation lies in its restricted interactivity and narrow task scope, making it insufficient for complex reasoning, planning, or rich object-agent interactions required in broader AI scenarios.

\textbf{ProcTHOR} \cite{procthor} introduces a procedural generation framework for AI2-THOR, enabling large-scale creation of diverse indoor scenes via compositional layouts. This approach improves generalization by increasing environmental diversity, yet remains confined to constrained household domains. Moreover, its generated scenes are primarily geometric variants rather than semantically grounded or multimodal in origin.

In summary, existing simulators including VirtualHome, AI2-THOR, OmniGibson, Habitat, and ProcTHOR have significantly advanced embodied AI research by offering specialized environments tailored to specific domains. Yet, their scope and interactivity constraints underscore the need for a more comprehensive simulator capable of addressing diverse research demands simultaneously. VirtualEnv responds to these limitations by offering a unified, highly interactive, and scalable urban simulation platform that incorporates vision-language models (VLMs) and large language models (LLMs) to procedurally generate semantically grounded environments and tasks. This enables the study of advanced cognitive behaviors, adaptive decision-making, and systematic benchmarking of LLMs in complex, dynamic scenarios.

%% file: text/environment.tex
\section{VirtualEnv}

\begin{figure*}[t]
    \hfill
    \includegraphics[width=\linewidth]{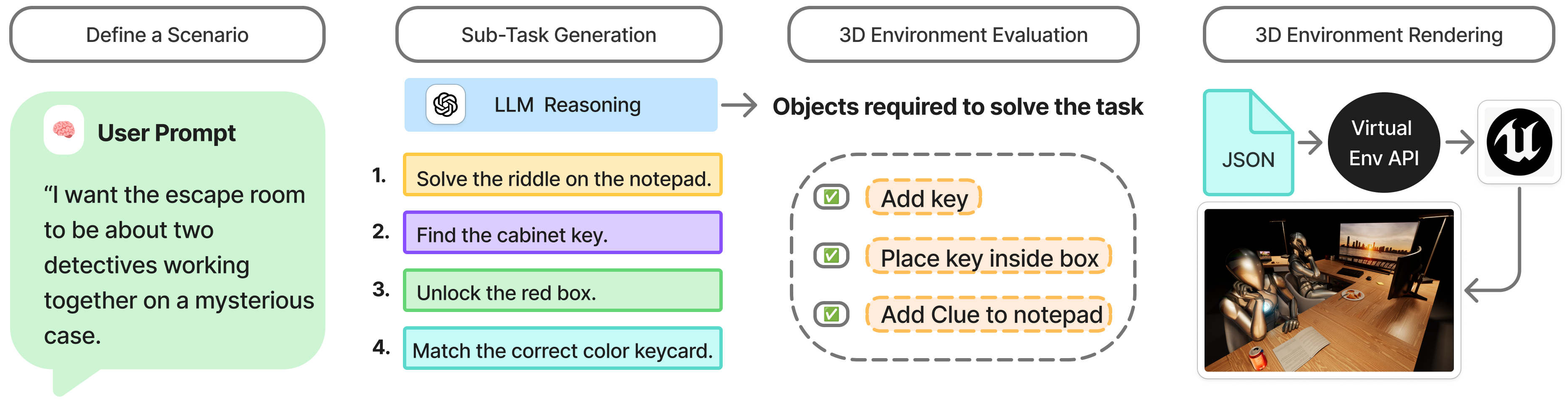}
     \caption{Language-based task and scenario generation in VirtualEnv. A user provides a natural language prompt describing a high-level scenario, which is parsed by a vLLM into sub-tasks and structured goals (e.g., solving riddles, unlocking containers). Based on the task requirements, VirtualEnv automatically evaluates which objects are needed, updates the scene graph accordingly, and renders the environment through the VirtualEnv API. This pipeline enables flexible and scalable creation of interactive, goal-driven scenarios without manual scripting.}
    
    \label{fig:taskgen_pipeline}
\end{figure*}

In this section, we introduce the main features of VirtualEnv, built on Unreal Engine 5 to support complex agent-object interactions and procedurally generated environments. VirtualEnv includes diverse indoor and outdoor settings, multimodal sensing capabilities, and high-resolution object models. The platform is tightly integrated with large language models (LLMs) through a lightweight Python API, enabling agents to interpret instructions, plan actions, and dynamically interact with the environment using language. We describe core components such as scene representation and interactive task design in Section~\ref{main_feat}, and demonstrate how VirtualEnv enables escape room–style tasks for evaluating problem-solving and emergent AI behaviors.

\subsection{Main Features of the VirtualEnv Simulator}
\label{main_feat}

Figure~\ref{fig:example} showcases VirtualEnv's three core capabilities: photorealistic indoor-outdoor environments, an extensive library of 20,000+ interactive assets, and precise control over humanoid agents. These features form the foundation for our comprehensive embodied AI research platform.

As shown in Table~\ref{tab:comparison}, VirtualEnv uniquely combines multi-agent support and language interaction capabilities while offering the most comprehensive environment type (3D-MIO) and the largest task library (140,000 tasks) among existing platforms. This combination enables complex scenarios that span both indoor and outdoor settings, supporting a broader range of embodied AI research than single-room (AI2Thor) or multi-room-only (Habitat 3.0, VirtualHome) alternatives.

\textbf{High-Fidelity Engine:} 
VirtualEnv provides a highly dynamic and interactive AI research environment. Built on Unreal Engine 5 \cite{unrealengine5}, it features diverse settings, including offices, retail venues, and urban streetscapes. At its core, VirtualEnv emphasizes realism and complex agent-environment interactions, using advanced rendering pipelines and procedural generation to enable boundless variations in physical layouts, object placements, and lighting conditions.

\textbf{Rich Object and Action Library:}
With over 20,000 distinct objects, VirtualEnv supports diverse real-world scenarios such as home furnishing, household activities, urban navigation, and multi-step decision-making. Each object is embedded with affordances—e.g., openable doors, movable furniture, graspable objects, and interactive appliances—allowing agents to perform fine-grained interactions. Many objects use photogrammetry scans for high-resolution modeling, ensuring realistic physics and visual accuracy. Unreal Engine's physics engine enables authentic object responses to interactions such as movement, deformation, and state transitions. Additionally, structured metadata allows AI agents to reason about object properties, enhancing their ability to learn real-world physical interactions.

\textbf{Multi-Modal Sensing and Observations:}
VirtualEnv offers rich multimodal sensing capabilities to support perception and decision-making in dynamic environments. Agents access RGB and depth sensors for photorealistic input and spatial understanding, as well as semantic segmentation for pixel-level object recognition. Panoramic top-down views further aid spatial reasoning and large-scale navigation. These modalities together enable agents to interpret and act within complex scenes with greater precision and adaptability.

\begin{figure*}[t]
    \hfill
    \includegraphics[width=\linewidth]{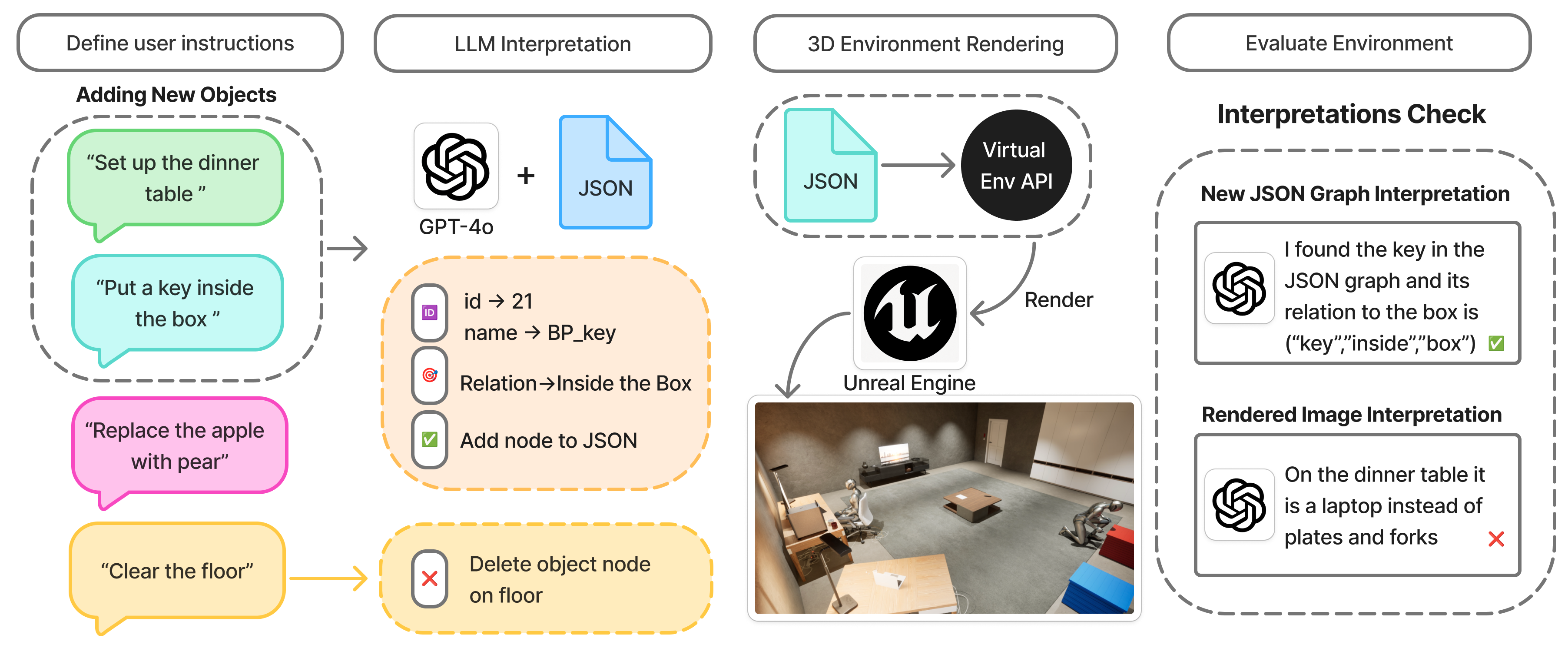}
    \caption{Interactive environment editing and interpretation validation in VirtualEnv. Users provide natural language instructions to modify the environment (e.g., adding, replacing, or removing objects). A vLLM interprets these instructions and updates the scene graph, which is then rendered using the VirtualEnv API. To ensure semantic alignment between the symbolic scene graph and the visual output, the system performs interpretation checks on both the JSON graph and the rendered image. This process enables reliable, language-driven environment manipulation and validation.}
    \label{fig:envgen_vllm}
\end{figure*}

\textbf{User-Friendly API and Language-Driven AI Agents:}
VirtualEnv natively supports the integration of large language models (LLMs) and vision-language models (VLMs), allowing AI agents to interact with the world through natural language. This enables flexible task execution, dynamic decision-making, and interactive environment control based on high-level instructions. Researchers can explore how AI models interpret, respond to and act on language commands in real time, supporting advances in LLM-based robotics and embodied language understanding.

\textbf{Scene Graph Representation:}
VirtualEnv organizes its environments using a scene graph, which encodes objects, agents, and spatial relationships in a hierarchical structure. This representation allows for efficient querying of environmental states, enabling agents to make informed decisions based on their surroundings. It also supports semantic reasoning, allowing agents to understand object affordances and spatial constraints, improving their ability to interact with the environment meaningfully. Additionally, the scene graph facilitates partial observations, making it possible to study agent behavior in scenarios where only limited information is available, which is crucial for research on planning under uncertainty. By providing a programmable scene graph API, VirtualEnv simplifies the process of creating and modifying scenes and tasks, offering researchers greater flexibility in designing interactive and adaptive simulation environments.

\textbf{Environment Construction and Asset Selection:}
All environments in \textbf{VirtualEnv} were constructed using a hybrid approach that combined manual scene authoring and procedural generation within Unreal Engine 5. We curated high-resolution 3D assets from the Unreal Engine Marketplace, selecting those with rich affordances (e.g., openable, graspable, movable) and clear semantic categories. These assets were chosen based on their relevance to embodied AI tasks—such as indoor navigation, object manipulation, and goal-directed behavior—and were organized into thematically consistent scenes (e.g., offices, kitchens, retail stores, and urban outdoor environments).

\textbf{Language Driven Task and Scenario Generation:}
VirtualEnv supports dynamic scenario generation through a language based interface. Users provide natural language prompts describing desired task setups, such as escape room challenges or collaborative problem solving scenarios. An LLM interprets the prompt and decomposes it into a sequence of subgoals or puzzles (e.g., "find the key," "solve a riddle," "unlock the box"). Based on these subgoals, the system identifies the required environmental components including objects, clues, and spatial arrangements, and automatically updates the scene graph to instantiate them. The environment is then rendered through the VirtualEnv API, resulting in a complete interactive scene where agents can be evaluated on their ability to complete the language defined tasks. This pipeline allows for scalable and diverse environment generation grounded in multimodal reasoning and task aware scene construction. (See Figure~\ref{fig:taskgen_pipeline})

\begin{figure*}[t]
\centering
\setlength{\tabcolsep}{0pt}
\renewcommand{\arraystretch}{1.0}
\begin{tabular}{c@{\hspace{1mm}}c@{\hspace{1mm}}c@{\hspace{1mm}}c@{\hspace{1mm}}c}
\begin{tabular}{@{}c@{}}
  \includegraphics[width=0.19\textwidth]{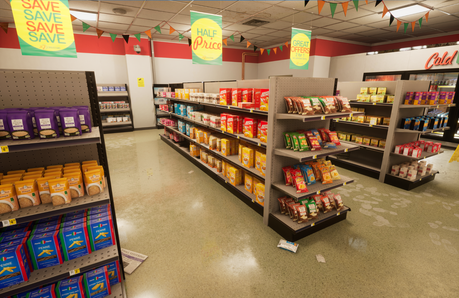}\\[-5pt]
  \includegraphics[width=0.19\textwidth]{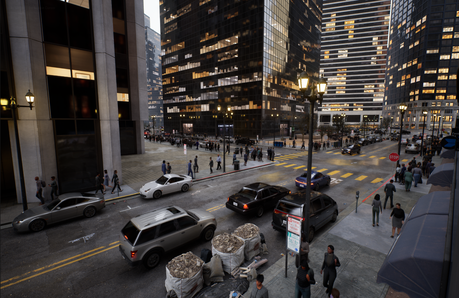}\\[-5pt]
  \includegraphics[width=0.19\textwidth]{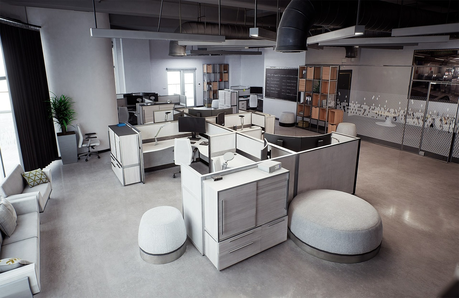}\\
  \textbf{VirtualEnv}\\
  \textbf{4.46 $\pm$ 1.02}
\end{tabular}
&
\begin{tabular}{@{}c@{}}
  \includegraphics[width=0.19\textwidth]{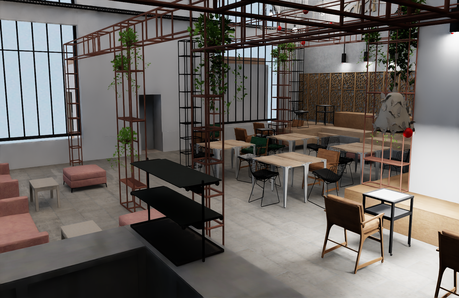}\\[-5pt]
  \includegraphics[width=0.19\textwidth]{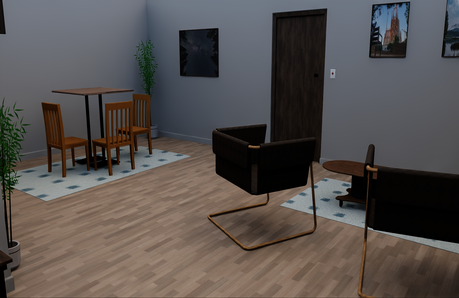}\\[-5pt]
  \includegraphics[width=0.19\textwidth]{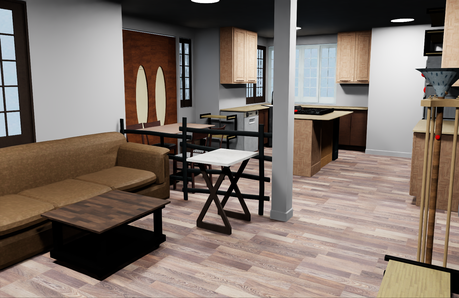}\\
  OmniGibson\\
  3.35 $\pm$ 1.05
\end{tabular}
&
\begin{tabular}{@{}c@{}}
  \includegraphics[width=0.19\textwidth]{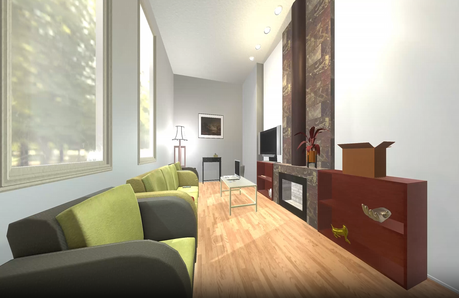}\\[-5pt]
  \includegraphics[width=0.19\textwidth]{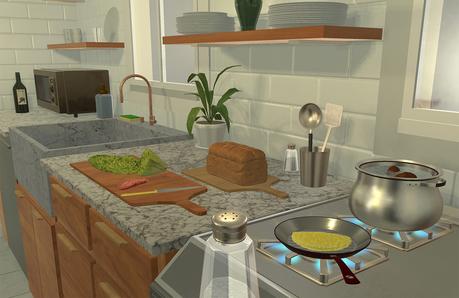}\\[-5pt]
  \includegraphics[width=0.19\textwidth]{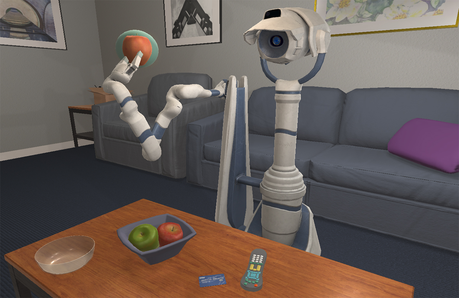}\\
  AI2THOR\\
  2.99 $\pm$ 1.08
\end{tabular}
&
\begin{tabular}{@{}c@{}}
  \includegraphics[width=0.19\textwidth]{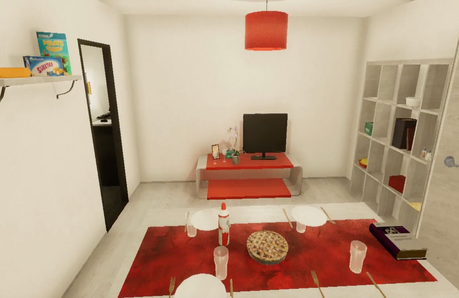}\\[-5pt]
  \includegraphics[width=0.19\textwidth]{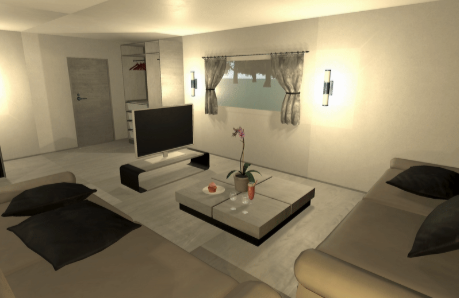}\\[-5pt]
  \includegraphics[width=0.19\textwidth]{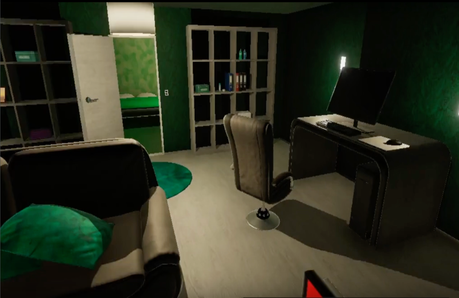}\\
  VirtualHome\\
  2.63 $\pm$ 1.24
\end{tabular}
&
\begin{tabular}{@{}c@{}}
  \includegraphics[width=0.19\textwidth]{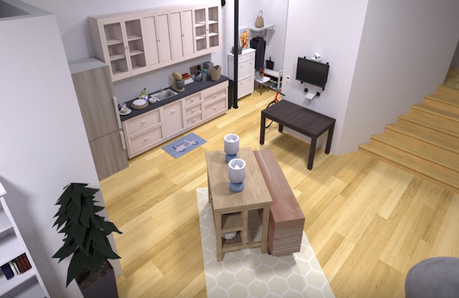}\\[-5pt]
  \includegraphics[width=0.19\textwidth]{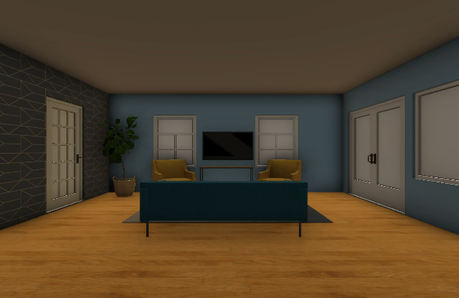}\\[-5pt]
  \includegraphics[width=0.19\textwidth]{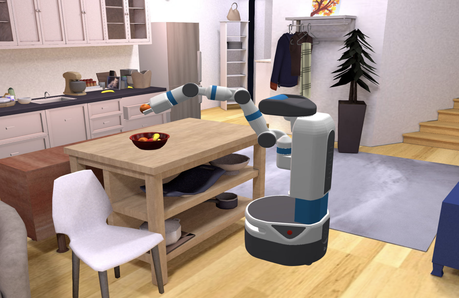}\\
  Habitat\\
  1.85 $\pm$ 1.09
\end{tabular}
\end{tabular}
\caption{Comparison of Visual Realism Rankings Across Platforms. A qualitative benchmarking study was conducted using a survey with 31 respondents. Participants ranked each platform based on visual realism, assigning a score from 5 (most realistic) to 1 (least realistic) in a label-blind test. We observe that the participants consider \textbf{VirtualEnv} to be significantly more visually realistic than all other environments.}
\label{fig:visual_realism}
\end{figure*}

\subsection{Escape Room Challenge Framework}
To evaluate higher-level reasoning in VirtualEnv, we introduce an Escape Room Challenge Framework. Unlike simple navigation or retrieval tasks, escape rooms blend puzzle-solving, object interactions, narrative clues, and occasional multi-agent coordination. Agents must discover clues in the environment, connect information across the scene, and solve an overarching puzzle, encouraging more flexible and deliberate behavior.

\textbf{Rationale and Design.}
Puzzle design in games provides a useful structure for planning and problem-solving because each puzzle is new and cannot be solved through memorization. Many puzzles also involve abstract or fictional objects, which require agents to adapt to unfamiliar situations. Incorporating these ideas into VirtualEnv encourages exploration, strategy refinement, and responsiveness to environmental feedback. Our framework integrates cognitive puzzles, interaction mechanics, and narrative hints, following the experience-pyramid model of \cite{Heikkinen2016DesigningAE}. Agents must navigate, manipulate objects, and make decisions as the puzzle unfolds.

\textbf{Levels of Complexity:}
We categorize our escape room challenges into four difficulty levels based on puzzle length and inter-clue dependencies, progressively increasing cognitive demands:

\begin{enumerate}
    \item \textbf{Level 1 - One Step Problem:} A single clue leads directly to the key, requiring minimal inference. Agents parse a textual hint and execute a basic action sequence to unlock the door.
    
    \item \textbf{Level 2 - Sequential Puzzles:} Agents must complete an intermediate task (e.g., arranging colored objects) to reveal the real clue, introducing multi-step reasoning.
    
    \item \textbf{Level 3 - Meta Clues:} Two parallel puzzles generate separate clues, both necessary for finding the key. Agents must integrate multiple information sources, reinforcing contextual reasoning.
    
    \item \textbf{Level 4 - Deceptive Clues:} Agents receive two clues, one accurate and one misleading. They must determine which clue is correct. These puzzles challenge critical thinking, error-checking, and contextual understanding.
\end{enumerate}

\textbf{Implications for AI Research:}
By embedding escape-room puzzles in VirtualEnv's realistic, dynamic settings, we bridge embodied AI and next-generation gaming applications. Success in these tasks signals advances in AI reasoning, contextual comprehension, and adaptive problem-solving. Additionally, our flexible puzzle design provides a framework to explore LLM fine-tuning, embodied AI integration, and generalization to novel tasks. As AI continues to evolve, these benchmarks will help assess how well AI agents adapt to unfamiliar, multi-step challenges in real-time interactive environments.

\begin{figure}[t]
  \centering
  \includegraphics[width=\linewidth]{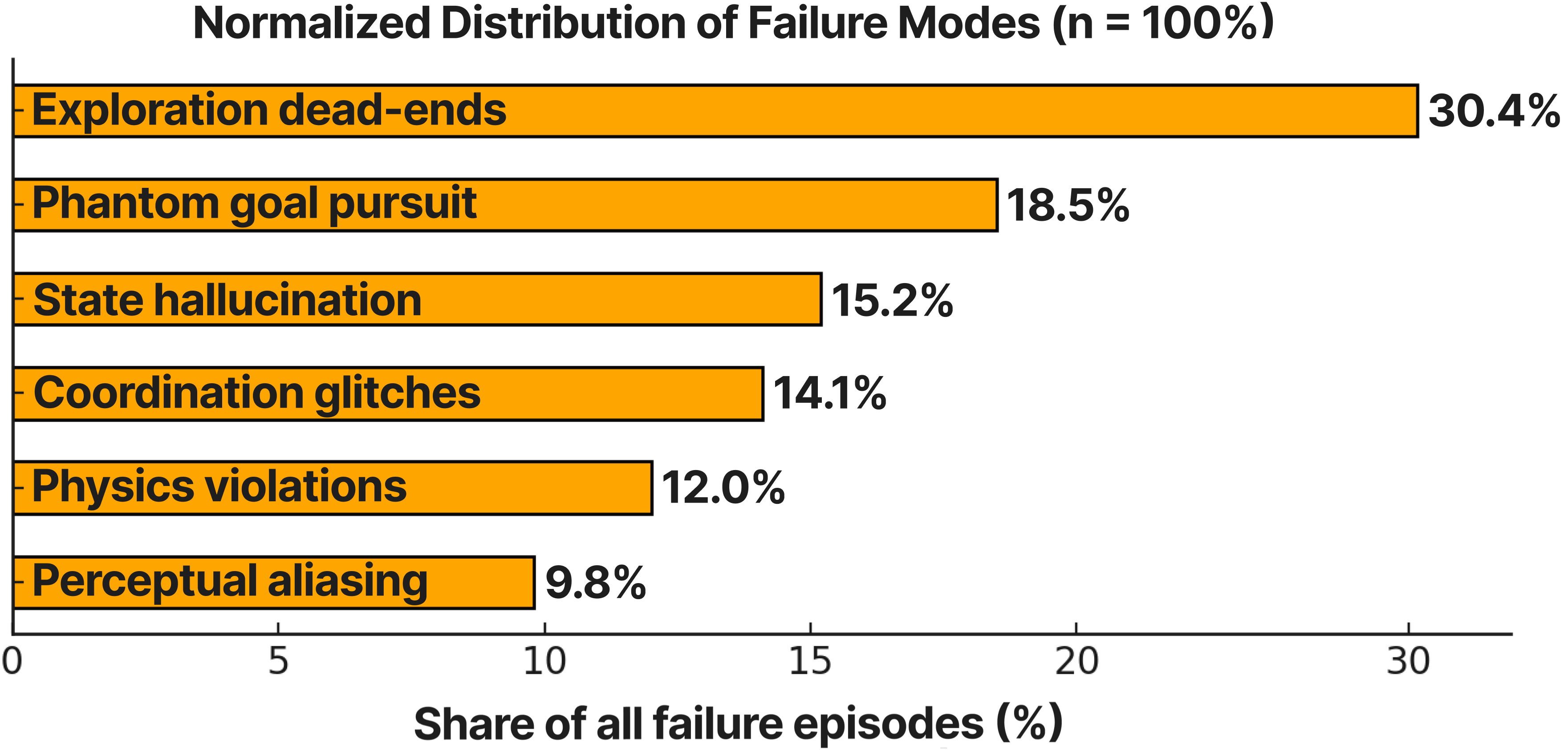}
  \caption{Distribution of Failure Modes in Embodied AI Tasks. Analysis of failure modes reveals six primary categories. }
  \label{fig:failure_modes_distribution}
\end{figure}

\subsection{Environment Modification with vLLM}
\label{sec:env_generation}

As illustrated in Figure~\ref{fig:envgen_vllm}, VirtualEnv uses a vision‑language model (vLLM) to modify an existing 3‑D scene in response to natural‑language commands. The model converts a prompt (e.g., “put a key inside the box”) into JSON‑encoded edits that specify the target objects, spatial relations, and placement rules. These edits are merged into the current observation graph and rendered in Unreal Engine 5. An interpretation check then compares the symbolic graph with the rendered view, flagging any mismatches between the intended and visualized states. This language‑guided editing pipeline enables rapid, precise adjustment of complex environments without manual intervention.

\begin{table*}[t]
  \centering
  \footnotesize
  \resizebox{\textwidth}{!}{%
    \begin{tabular}{l c c c c c c c c}
      \toprule
       & \multicolumn{4}{c}{\textbf{Reasoning LLMs}}
       & \multicolumn{4}{c}{\textbf{Non-Reasoning LLMs}} \\
      \cmidrule(lr){2-5}\cmidrule(lr){6-9}
      \textbf{Task}
        & Claude 3 Opus 
        & Gemini 2.5 Pro 
        & o3 
        & Grok 3 Think 
        & GPT-4o 
        & Llama 3.1 405B 
        & Qwen 2.5 Max 
        & Llama 4 (MoE) 
        \\
      \midrule
      Clean Floor (S) & $0.85\!\pm\!0.03$ & $0.83\!\pm\!0.04$ & $0.82\!\pm\!0.04$ & $0.76\!\pm\!0.05$
                      & $0.68\!\pm\!0.05$ & $0.62\!\pm\!0.06$ & $0.61\!\pm\!0.05$ & $0.60\!\pm\!0.06$ \\[7pt]
      Watch TV (S)    & $0.88\!\pm\!0.02$ & $0.86\!\pm\!0.03$ & $0.85\!\pm\!0.03$ & $0.80\!\pm\!0.04$
                      & $0.72\!\pm\!0.04$ & $0.66\!\pm\!0.05$ & $0.65\!\pm\!0.05$ & $0.64\!\pm\!0.05$ \\[7pt]
      Find Object (S) & $0.70\!\pm\!0.05$ & $0.68\!\pm\!0.05$ & $0.64\!\pm\!0.06$ & $0.60\!\pm\!0.06$
                      & $0.48\!\pm\!0.06$ & $0.46\!\pm\!0.07$ & $0.45\!\pm\!0.07$ & $0.40\!\pm\!0.08$ \\[7pt]
      Prepare Food (M)& $0.92\!\pm\!0.03$ & $0.90\!\pm\!0.03$ & $0.88\!\pm\!0.04$ & $0.84\!\pm\!0.04$
                      & $0.75\!\pm\!0.04$ & $0.70\!\pm\!0.05$ & $0.69\!\pm\!0.05$ & $0.68\!\pm\!0.05$ \\[7pt]
      Clean Room (M)  & $0.93\!\pm\!0.02$ & $0.92\!\pm\!0.03$ & $0.90\!\pm\!0.03$ & $0.86\!\pm\!0.04$
                      & $0.78\!\pm\!0.04$ & $0.74\!\pm\!0.05$ & $0.73\!\pm\!0.05$ & $0.72\!\pm\!0.05$ \\
      \bottomrule
    \end{tabular}%
  }
  \caption{Performance Comparison of Reasoning vs. Non-Reasoning LLMs on Embodied Tasks. Success rates ($\pm$ 1 Standard Deviation) across five benchmark tasks, including both single-agent (S) and multi-agent (M) scenarios. Reasoning LLMs consistently outperform their non-reasoning counterparts, with the performance gap being most pronounced in complex tasks like \textit{Find Object} and \textit{Prepare Food}. Multi-agent tasks generally show higher success rates, demonstrating the benefits of collaborative planning.}
  \label{tab:model_performance}
\end{table*}

%% file: text/experiments.tex
\section{Experiments}
\label{sec:experiments}

Our experiments begin with a comprehensive evaluation of VirtualEnv's visual fidelity through a qualitative benchmarking study, comparing it against leading simulation platforms in the field. We then assess the effectiveness of LLM-based planners within this high-fidelity environment, focusing on their ability to make structured decisions, coordinate tasks, and adapt to dynamic environments in both single-agent and multi-agent scenarios.

To achieve this, we design controlled experiments where agents operate in partially observable environments. The single-agent experiments evaluate fundamental skills such as navigation, object retrieval, and environmental manipulation, while the multi-agent experiments examine how collaborative planning improves task efficiency in scenarios requiring synchronized actions and division of labor.

To contextualize our findings, we benchmark agent performance within VirtualEnv, evaluating generalization across different task complexities. We assess each planner's ability to adapt to new tasks, using structured comparisons to measure how well agents scale to increasing environmental challenges, including both routine household tasks and complex puzzle-solving scenarios.

\subsection{Visual Realism}
To evaluate the visual realism of VirtualEnv in comparison to existing simulation platforms, we conducted a qualitative benchmarking study. Participants (N=31) were asked to rank multiple platforms (VirtualEnv, OmniGibson, AI2THOR, VirtualHome, and Habitat) based on visual realism through a label-blind survey. Each platform was rated from 5 (most realistic) to 1 (least realistic). As shown in Figure~\ref{fig:visual_realism}, VirtualEnv achieved a significantly higher realism score (4.46 ± 1.02) compared to other platforms, clearly demonstrating its advantage in generating visually realistic environments for embodied AI tasks.

\subsection{Baseline Model Performance Analysis}
\label{sec:quant_results}

\paragraph{``Reasoning'' vs.\ ``Non-Reasoning'' LLMs.}
As shown in Table \ref{tab:model_performance}, our experiments compare four LLM variants with chain-of-thought capabilities against their base models across five distinct tasks. The chain-of-thought models show an average improvement of 11\% in task completion rates, with particularly strong gains in complex, multi-step activities like \textit{Find Object} and \textit{Prepare Food}. These improvements stem from the models' enhanced ability to break down tasks into logical steps and maintain context throughout execution. The performance is also more consistent, with standard deviations below 0.05 for routine tasks, suggesting that structured reasoning leads to both higher success rates and more reliable performance.

\paragraph{Task-specific difficulty profile.}
Success rates vary significantly across tasks. While \textit{Watch TV} achieves high performance (above \(0.85\)) for state-of-the-art models, the open-ended search required in \textit{Find Object} reduces performance by up to \(25\) percentage points and nearly doubles variance (\(\sigma = 0.06\!-\!0.08\)). This suggests that partial observability remains the primary challenge, even with photorealistic rendering and large language priors. To address this limitation, we propose augmenting the planner with explicit spatial memory or learned exploration heuristics.

\subsection{Collaborative Planning Analysis}
\label{sec:collab_results}
Building on our single-agent findings, we observe that multi-agent collaboration consistently improves performance. For instance, on \textit{Prepare Food}, Claude 3 Opus's success rate increases from \(0.88\) to \(0.92\), while GPT-4o improves from \(0.68\) to \(0.75\). Analysis of replay logs reveals that this improvement stems from effective task allocation. For example, one agent would handle utensil retrieval while another manages appliance operation, thereby reducing action horizons and minimizing occlusion-related uncertainty.

\subsection{Failure Modes}
\label{sec:failure_modes}

Our analysis identifies six key failure modes in embodied AI tasks, with their distribution shown in Figure~\ref{fig:failure_modes_distribution}. The most common failure (30.4\%) occurs when agents get stuck in exploration loops, repeatedly visiting the same rooms without finding their target. This happens because agents lack effective strategies for systematically exploring unseen areas. The second most frequent issue (18.5\%) involves agents pursuing non-existent objects, often because their planning process loses track of what's actually available in the environment. Other significant failures include: agents incorrectly assuming object states (15.2\%), multi-agent coordination problems (14.1\%), physically impossible action sequences (12.0\%), and confusion between similar-looking objects (9.8\%). These failure modes have clear implications for improving embodied AI systems. The top three categories, exploration loops, phantom goals, and state tracking errors account for nearly two-thirds of all failures. Addressing these core issues could potentially improve overall task success rates by 7.4\%, bringing the best-performing models closer to human-level performance on routine tasks.

%% file: text/conclusion.tex
\section{Conclusion}
VirtualEnv is a next-generation Unreal Engine 5 simulation platform for embodied AI and language-driven interaction. It offers scalable, richly interactive environments where agents perform complex tasks such as navigation, multi-step manipulation, collaboration, and goal-directed planning. Powered by vision-language models, it supports procedural scenario generation, language-guided scene editing, and task planning in dynamic environments, providing improved interactivity, diversity, and visual fidelity over existing platforms and a strong foundation for future work at the intersection of AI and simulation.

%% file: text/acknowledgments.tex
\section{Acknowledgments}
    We would like to thank Xavier Puig for his helpful insight and advice for creating a virtual environment designed for Embodied AI.